%% file: main.tex
\documentclass[conference]{IEEEtran}
\IEEEoverridecommandlockouts

\usepackage{cite}
\usepackage{amsmath,amssymb,amsfonts}
\usepackage{algorithmic}
\usepackage{graphicx}
\usepackage{textcomp}
\usepackage{xcolor}
\def\BibTeX{{\rm B\kern-.05em{\sc i\kern-.025em b}\kern-.08em
    T\kern-.1667em\lower.7ex\hbox{E}\kern-.125emX}}

\usepackage[ngerman, english]{babel}
\usepackage{paralist, tabularx}
\usepackage{graphicx}
\usepackage{amsmath}
\usepackage{multirow}
\usepackage{subcaption}
\usepackage{algorithm}
\usepackage{listings}
\usepackage{booktabs}
\usepackage{amssymb}
\usepackage{url}
\begin{document}

\title{
Squat: Quant Small Language Models on the Edge
}

\author{
Xuan Shen$^1$, Peiyan Dong$^2$, Zhenglun Kong$^3$, Yifan Gong$^1$, Changdi Yang$^1$, Zhaoyang Han$^1$, 
\\
Yanyue Xie$^1$, Lei Lu$^1$, Cheng Lyu$^1$, Chao Wu$^1$, Yanzhi Wang$^1$ and Pu Zhao$^1$
\\
$^1$Northeastern University, $^2$MIT, $^3$Harvard University
}

\maketitle

\input{sections/0_abstract}

\input{sections/1_introduction}

\input{sections/2_related_work}

\input{sections/3_analysis}

\input{sections/4_method}
\input{sections/5_results}

\input{sections/6_conclusion}

\bibliographystyle{abbrv}
\bibliography{reference}


\end{document}

%% file: sections/0_abstract.tex
\begin{abstract}

A growing trend has emerged in designing high-quality Small Language Models (SLMs) with a few million parameters. 
This trend is driven by the increasing concerns over cloud costs, privacy, and latency.
Considering that full parameter training is feasible for SLMs on mobile devices, Quantization-Aware Training (QAT) is employed to improve efficiency by reducing computational overhead and memory footprint.
However, previous QAT works adopt fine-grained quantization methods to compress models with billions of parameters on GPUs, incompatible with current commodity hardware, such as mobile and edge devices, which relies on Single Instruction Multiple Data (SIMD) instructions.
Thus, the generalization of these methods to SLMs on mobile devices is limited.
In this paper, we propose Squat method, an effective QAT framework with deployable quantization for SLMs on mobile devices.
Specifically, we propose entropy-guided and distribution-aligned distillation to mitigate the distortion of attention information from quantization.
Besides, we employ sub-8-bit token adaptive quantization, assigning varying bit widths to different tokens based on their importance.
Furthermore, we develop a SIMD-based Multi-Kernel Mixed-Precision (MKMP) multiplier to support sub-8-bit mixed-precision MAC on mobile devices.
Our extensive experiments verify the substantial improvements of our method compared to other QAT methods across various datasets.
Furthermore, we achieve an on-device speedup of up to 2.37$\times$ compared with its FP16 counterparts, signaling a great advancement. 
\textcolor{blue}{Code: \url{https://github.com/shawnricecake/squant}}

\end{abstract}

\begin{IEEEkeywords}
Quantization-Aware Training, Mobile, Small Language Models
\end{IEEEkeywords}

%% file: sections/1_introduction.tex
\section{Introduction}

Large Language Models (LLMs)~\cite{zhang2022opt, radford2019language, brown2020language, gpt3, touvron2023llama} have emerged as the dominant force in the field of Natural Language Processing (NLP).
These models are increasingly integrated into various applications on daily life devices to enhance task performance and user experience. 
With growing concerns about the cost and latency of deploying LLMs in the cloud, the trend is increasingly shifting towards deploying Small Language Models (SLMs) on edge platforms such as mobile devices.
Recently, MobileLLM~\cite{liu2024mobilellm}, SmolLM2~\cite{allal2025smollm2}, BabyLlama~\cite{timiryasov2023babyllama, tastet2024babyllama2} and other works~\cite{van2024slmsurvey, shen2025numerical, zhao2024fully} have prioritized the design of high-quality SLMs with fewer than one billion parameters, making such lightweight models a viable option for mobile deployment.
Meanwhile, given the widespread support for quantized computation by Single Instruction Multiple Data (SIMD) mechanisms on mobile and edge devices~\cite{ashfaq2022accelerating, shen2024hotaq, shen2025lazydit, shen2024search, shen2025sparse, shen2025quartdepth, shen2025draft, shen2025fastcar}, model quantization emerges as a promising approach to accelerate SLMs on these platforms. 
Existing methods using Post-Training Quantization (PTQ) often suffer significant accuracy drops in sub-8-bit settings.
Some PTQ methods like GPTQ~\cite{frantar2022gptq}, SqueezeLLM~\cite{kim2023squeezellm}, and AWQ~\cite{lin2023awq} only adopt the weight-only quantization, which leave the activations in float16 and can not fully exploit efficient integer computation on edge devices
~\cite{frantar2022gptq, lin2023awq, shen2024agile, xiao2023smoothquant}.
Other PTQ methods, including Agile-Quant~\cite{shen2024agile}, SmoothQuant~\cite{xiao2023smoothquant}, and ZeroQuant~\cite{yao2022zeroquant, wu2023zeroquantfp}, quantize both weights and activations. However, their performance drops significantly when the activation bit width is reduced to 4-bit.
In contrast, Quantization-Aware Training (QAT) allows fine-tuning to reduce quantization error, and models with both weights and activations quantized can take full advantage of fast integer matrix multiplication on edge hardware.
Meanwhile, for SLMs, full parameter training is feasible, making QAT a practical and effective option. This enables us to jointly optimize both weights and activations during training, leading to better performance at lower bit widths—especially when targeting 4-bit deployment.

Recent QAT works~\cite{kim2024tsld, liu2023llmqat, chen2024efficientqat} employ channel-wise or token-wise (fine-grained) quantization for weights and activations with sub-8-bit precision, resulting in multiple scaling factors for a single matrix. 
However, such fine-grained quantization methods are incompatible with computation kernels in general mobile devices and edge processors, 
leading to significant computation overhead.
Specifically, standard SIMD-based libraries~\cite{jacob2017gemmlowp,dukhan2018qnnpack} do not support running quantized networks with sub-8-bit precision, and the general matrix multiply (GeMM) kernel in SIMD cannot handle matrix multiplications (MAC) with multiple scaling factors for integer operations~\cite{ashfaq2022accelerating}.
In conclusion, to effectively harness the power of SLMs on mobile devices, it is necessary to address the challenges of quantization algorithm design and corresponding efficient hardware implementation.

Therefore, to build efficient and accurate SLMs on mobile devices, we present a novel framework called \textbf{Squat} in this paper.
Specifically, we first identify that, for the self-attention module, the information distortion brought by quantizing queries and keys is the most critical factor leading to information loss.
We then introduce an entropy-guided optimization method to mitigate the loss by maximizing the information entropy.
Simultaneously, we align the distributions of the quantized attention maps to the FP16 one to minimize the difference.
Furthermore, the sub-8-bit token adaptive quantization is implemented, which assigns varying bit widths to different tokens based on their informativeness, thereby further reducing redundancy.
Moreover, 
we recompile the existing INT8 multiplier and develop a SIMD-based Multi-Kernel Mixed-Precision (MKMP) multiplier for the proposed token adaptive quantization method. 
Finally, Squat framework can accelerate the inference of mixed-precision SLMs with sub-8-bit quantization  on mobile devices and edge processors.
In our experiments, we uniformly quantize the weights with 4 bits or 8 bits, and adaptively quantize the activations with 4 bits or 8 bits.
The results show that our method can achieve better task performance 
than other coarse-grained QAT methods.
Moreover, our results demonstrate that our proposed adaptive quantization approach yields superior performance with a mixed strategy compared to a uniform strategy.
For instance, a combination of half 4-bit and half 8-bit quantization outperforms uniform 6-bit quantization.
Furthermore, our quantized models with the proposed MKMP multiplier can achieve a practical speedup of up to 2.37$\times$ on mobile devices.
Our contributions are summarized below:
\begin{itemize}[]
    
    \item \textbf{1.} 
    We design the entropy-guided and distribution-aligned QAT method to mitigate information distortion brought by quantization.

    \item \textbf{2.} We design the token importance-aware adaptive quantization method for activations (i.e., tokens), further reducing model redundancy and outperforming bit-equivalent uniform quantization.

    \item \textbf{3.} We develop a SIMD-based MKMP multiplier for the acceleration of sub-8-bit mixed-precision LLM inference on mobile devices. 

    \item \textbf{4.} We achieve better task performance than other QAT methods
    with an on-device speedup of up to 2.37$\times$.
    
\end{itemize}

%% file: sections/2_related_work.tex
\section{Background and Related Works}
\label{sec:related_works}

\subsection{Efficient Design of SLMs}
Recently, a growing trend~\cite{van2024slmsurvey} is the use of SLMs for specialized tasks on mobile devices. These models, such as MiniCPM~\cite{hu2024minicpmunveilingpotentialsmall}, Octopus~\cite{chen2024octopusv2ondevicelanguage}, SmolLM2~\cite{allal2025smollm2}, and MobileLLM~\cite{liu2024mobilellm}, have shown that they can perform effectively on general tasks while being more suitable for deployment on mobile platforms. 
To facilitate SLM developments, quantization has emerged as a powerful technique to reduce the computational and memory demands with low-bit format.
Current quantization methods can be broadly categorized into PTQ and QAT.
Previous weight-only quantization methods, such as GPTQ~\cite{frantar2022gptq}, SqueezeLLM~\cite{kim2023squeezellm}, and AWQ~\cite{lin2023awq}, focus on optimizing quantized weights while leaving activations unquantized. This approach fails to exploit efficient integer computation capabilities on edge devices. 
PTQ works, such as SmoothQuant~\cite{xiao2023smoothquant}, ZeroQuant~\cite{yao2022zeroquant}, ZeroQuant-FP~\cite{wu2023zeroquantfp}, and Agile-Quant~\cite{shen2024agile} can achieve multiple different quantization configurations for both weights and activations. 
On the other hand, QAT methods, like LLM-QAT~\cite{liu2023llmqat}, TSLD~\cite{kim2024tsld}, EfficientQAT~\cite{chen2024efficientqat}, and others~\cite{tao2022compression} use fine-tuning with fine-grained quantization (e.g., channel-wise and token-wise ) to recover task performance.
However, these fine-grained quantization techniques are difficult to be deployed on mainstream edge processors, making them less practical for mobile devices.

\subsection{Hardware Implementation for Quantization}

Low-precision linear algebra kernels are designed to maximize computing throughput by adapting existing wider bit-width kernels to handle lower-precision operands.
This approach improves performance based on lower-precision SIMD instructions (e.g., vmlaq s8() in ARMv8 ISA), which process more elements in parallel than higher-precision instructions (e.g., vmlaq f32()).
State-of-the-art (SOTA) low-precision linear algebra kernels, such as Google’s GEMMLOWP~\cite{jacob2017gemmlowp} and Meta’s QNNPACK~\cite{dukhan2018qnnpack}, 
can greatly enhance model efficiency with 8-bit quantization (W8A8) scenarios, e.g., on a 64-bit ARM Cortex-A72 CPU.
However, two critical challenges have yet to be addressed to achieve SOTA complex quantization frameworks.
(i) More aggressive sub-8-bit quantization does not offer additional performance benefits, as commercial  CPUs/GPUs only support SIMD operations with 8-bit or higher precision. 
Thus, low-precision kernels simply zero-extend sub-8-bit operands to byte boundaries, treating them as 8-bit operands.
(ii) Quantization process is executed with a SIMD-based GeMM engine that supports layer-wise (coarse-grained) quantization with a single scaling factor per matrix, which means the fine-grained quantization methods adopted by SOTA methods are limited on the edge.

%% file: sections/3_analysis.tex

\section{Analysis}
\label{sec:motivations}

To identify the bottlenecks of layer-wise QAT, we analyze performance deterioration when quantizing each part of the model. Additionally, we assess token importance using the attention map, specifically examining its relationship with the initial token of each head.

\begin{figure}[t]
    \centering
    \includegraphics[width=1.0\linewidth]{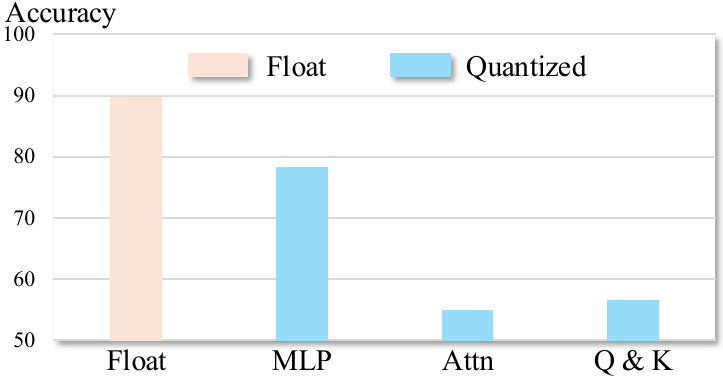}
  \caption{
    Accuracy analysis of LLaMA-58M on the Anaphor Agr. subdataset of BLiMP with different quantized modules.
    }
    \label{fig:analysis_accuracy}
\end{figure}

\begin{figure}[t]
\vspace{-3mm}
    \centering
  \includegraphics[width=1.0\linewidth]{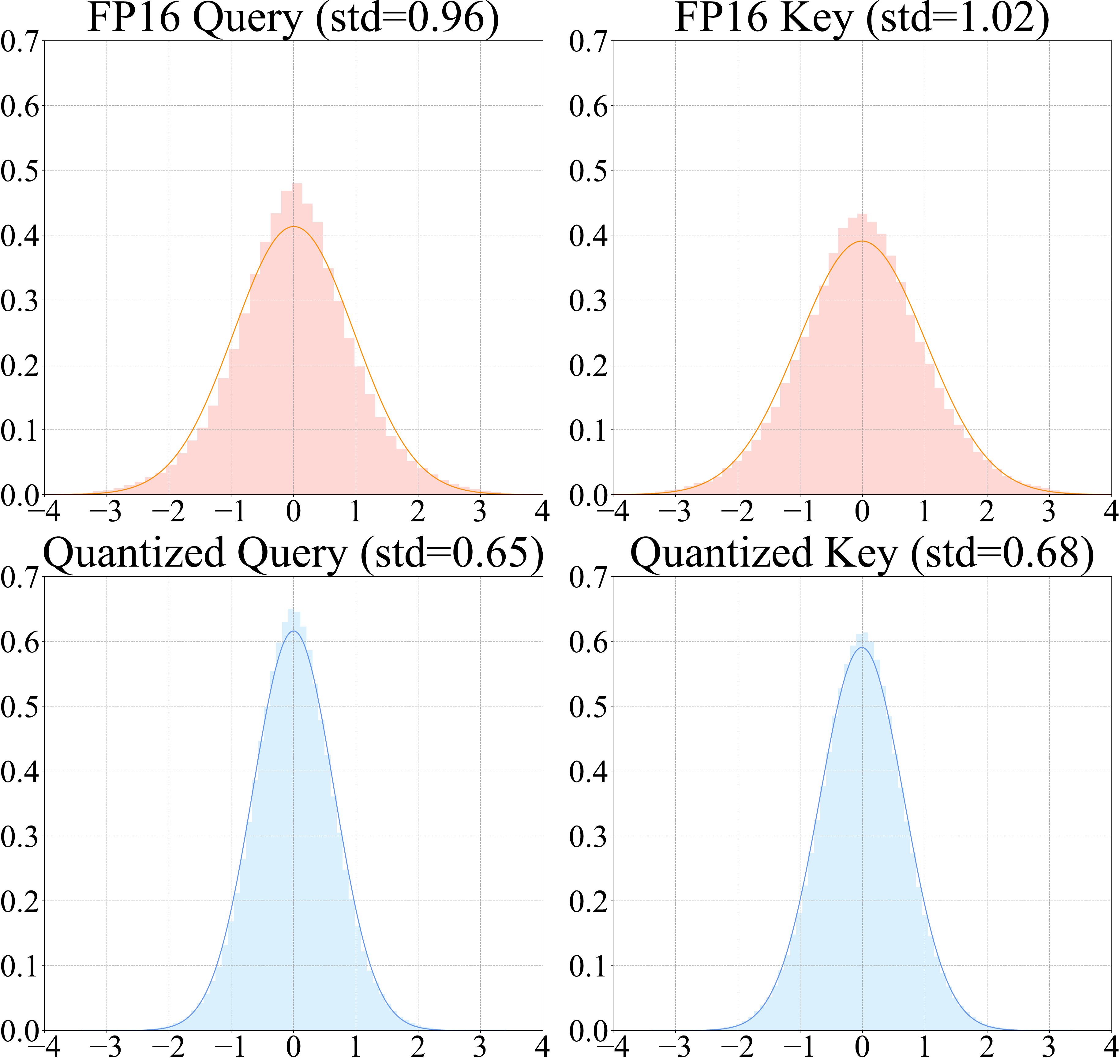}
  \caption{
  Distributions of query and key at the last layer of FP16 and quantized LLaMA-58M. 
  The main difference is from the variance of the distribution.
  }
  \label{fig:quant_query_key_distribution}
\end{figure}

\begin{figure}
    \centering
    \includegraphics[width=1.0\linewidth]{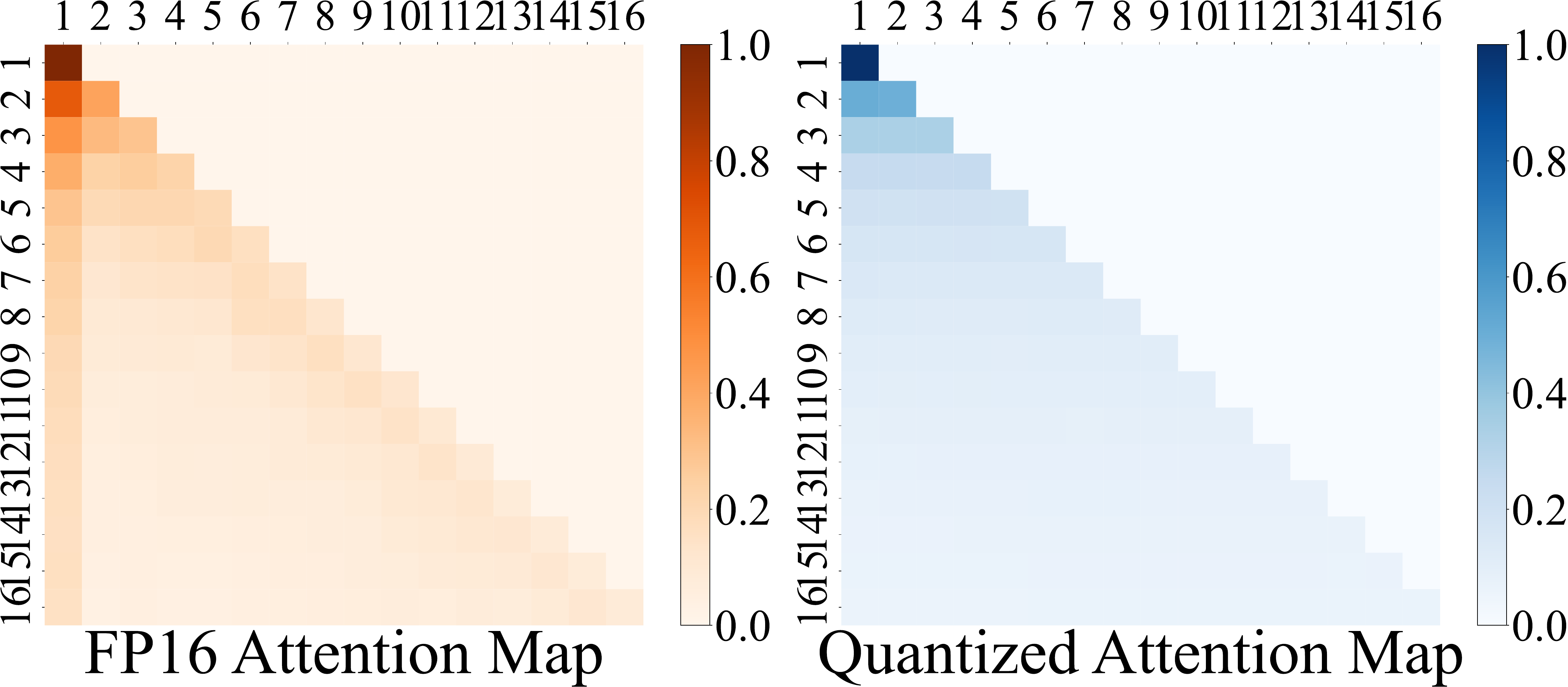}
  \caption{
  Averaged attention maps through all heads at the last layer of the FP16 and quantized models.
  The main difference is from the first column of the heatmap.
  }
  \label{fig:quant_quantized_fp16_attentionmap}
\end{figure}

\subsection{Quantized Self-Attention Module} 
We first analyze the degradation on task performance by applying layer-wise quantization (i.e., coarse-grained quantization) to various components of the model without fine-tuning.
In Figure~\ref{fig:analysis_accuracy}, the MLP module, the entire self-attention module, and specific parts of the self-attention module (query and key) are quantized, respectively.   Other parts of the model remain as FP16.
As shown in Figure~\ref{fig:analysis_accuracy}, we identify that the quantized query and key are the main factors that lead to the substantial performance drop, as the performance drop caused by query and key nearly mirrors the loss from quantizing the entire self-attention module.
We further visualize the distributions of quantized queries and keys, which approximate Gaussian distributions~\cite{qin2022bibert, shen2024agile}. As shown in Figure~\ref{fig:quant_query_key_distribution}, there is a significant difference in distributions between the quantized and FP16 versions for either  the query or key (e.g., different variance). 
The distribution difference results in an entropy discrepancy, inevitably leading to a deterioration in the attention module's representational capability.

\subsection{Token Importance} 

In LLMs, a unique initial token is placed at the start of the input sequence to define token positions, visible to all subsequent tokens due to autoregressive language modeling.
Removing interactions between the initial token and other ones can fundamentally alter the model's output.
However, as shown in Figure~\ref{fig:quant_quantized_fp16_attentionmap}, a distinct column pattern associated with the initial token in the FP16 version disappears in the quantized version.
Besides, we highlight that the initial token remains critical for assessing token importance, which is essential for further optimizations like token pruning~\cite{kim2024token,dong2023heatvit,kong2022spvit,shen2024agile}.
In generative models, the self-attention mechanism limits each token's interactions to those preceding it, and the initial token encapsulates the informational content of each generated token.
Thus, we evaluate token importance based on each token’s average attentivity to the initial token with all heads (i.e., the first column of attention map).

%% file: sections/4_method.tex
\section{Methodology}

In this section, we first provide the preliminary of QAT.
Then, we propose the design of entropy loss and distribution loss.
We further introduce token adaptive quantization method based on token importance.
We also design the MKMP multiplier for adaptive quantization deployment on mobile devices.

\subsection{Preliminary}

For layer-wise QAT, we adopt the symmetric quantization for both weights $\textbf{w}$ and activations $\textbf{x}$ as follows,
\begin{equation} 
    \mathcal{Q} (\textbf{w}) = \lfloor \text{CLIP}( \frac{\textbf{w}}{\alpha_{\textbf{w}}} , -2^{b_{\textbf{w}}-1}, 2^{b_{\textbf{w}}-1}-1)  \rceil ;
\end{equation}
\begin{equation}
    \hat{\textbf{w}} = \mathcal{Q} (\textbf{w}) \cdot \alpha_{\textbf{w}};
\end{equation}
\begin{equation} 
    \mathcal{Q} (\textbf{x}) = \lfloor \text{CLIP}( \frac{\textbf{x}}{\alpha_{\textbf{x}}} , -2^{b_{\textbf{x}}-1}, 2^{b_{\textbf{x}}-1}-1)  \rceil ;
\end{equation}
\begin{equation}
    \hat{\textbf{x}} = \mathcal{Q} (\textbf{x}) \cdot \alpha_{\textbf{x}},
\end{equation}
where $\mathcal{Q}(\cdot)$ denotes the quantization function.
CLIP$(x, r_1, r_2)$ returns $x$ with values in the range from $r_1$ to $r_2$ through clipping with the lower bound $r_1$  and upper bound $r_2$.
$\lfloor \cdot \rceil $ represents rounding to the nearest integer.
$\textbf x$ is the activations and $\textbf w$ means the weights.
$\hat{\textbf{x}}$ and $\hat{\textbf{w}}$ denote the dequantized activations and weights with scaling factor $\alpha$.
$b_{\textbf x}$ and $b_{\textbf w}$ denote bit width for activations and weights.

For the forward propagation, the linear projection can be calculated as follows:
\begin{equation}
\begin{aligned} 
    \mathcal{F}_{Linear} (\textbf{x}, \textbf{w}) &= \hat{\textbf{x}} \times \hat{\textbf{w}} \\ 
    &= \alpha_{\textbf{x}} \alpha_{\textbf{w}} \left[  \mathcal{Q}(\textbf{x}) \times  \mathcal{Q}(\textbf{w})  \right],
\end{aligned}
\end{equation}
where the $\mathcal{F}_{Linear}$ denotes the normal matrix multiplication.
For backward propagation, gradients are computed as follows:
\begin{equation}
\begin{aligned}
    \frac{\partial \mathcal{J}}{\partial \textbf{x}} &= \frac{\partial \mathcal{J}}{\partial \hat{\textbf{x}}} \frac{\partial \hat{\textbf{x}}}{\partial \textbf{x}} \\
    &= 
    \left\{
        \begin{array}{ll}
            \frac{\partial \mathcal{J}}{\partial \hat{\textbf{x}}}, & \textbf{x} \in \left[ -2^{b_{\textbf{x}}-1}, 2^{b_{\textbf{x}}-1}-1 \right], \\
            0, & \text{otherwise}.
        \end{array}
    \right.
\end{aligned}
\end{equation}
\begin{equation}
\begin{aligned}
    \frac{\partial \mathcal{J}}{\partial \textbf{w}} 
    &= \frac{\partial \mathcal{J}}{\partial \textbf{x}} 
       \frac{\partial \textbf{x}}{\partial \hat{\textbf{w}}} 
       \frac{\partial \hat{\textbf{w}}}{\partial \textbf{w}} \\
    &=
    \left\{
        \begin{array}{ll}
            \frac{\partial \mathcal{J}}{\partial \textbf{x}} 
            \frac{\partial \textbf{x}}{\partial \hat{\textbf{w}}}, 
            & \textbf{w} \in \left[ -2^{b_{\textbf{w}}-1}, 2^{b_{\textbf{w}}-1}-1 \right], \\
            0, & \text{otherwise}.
        \end{array}
    \right.
\end{aligned}
\end{equation}
where $\mathcal{J}$ denotes the loss function, and straight-through estimator (STE)~\cite{bengio2013estimating} is adopted to retain derivation of gradients.

\begin{figure*}[t]
  \centering
  \includegraphics[width=1.0\linewidth]{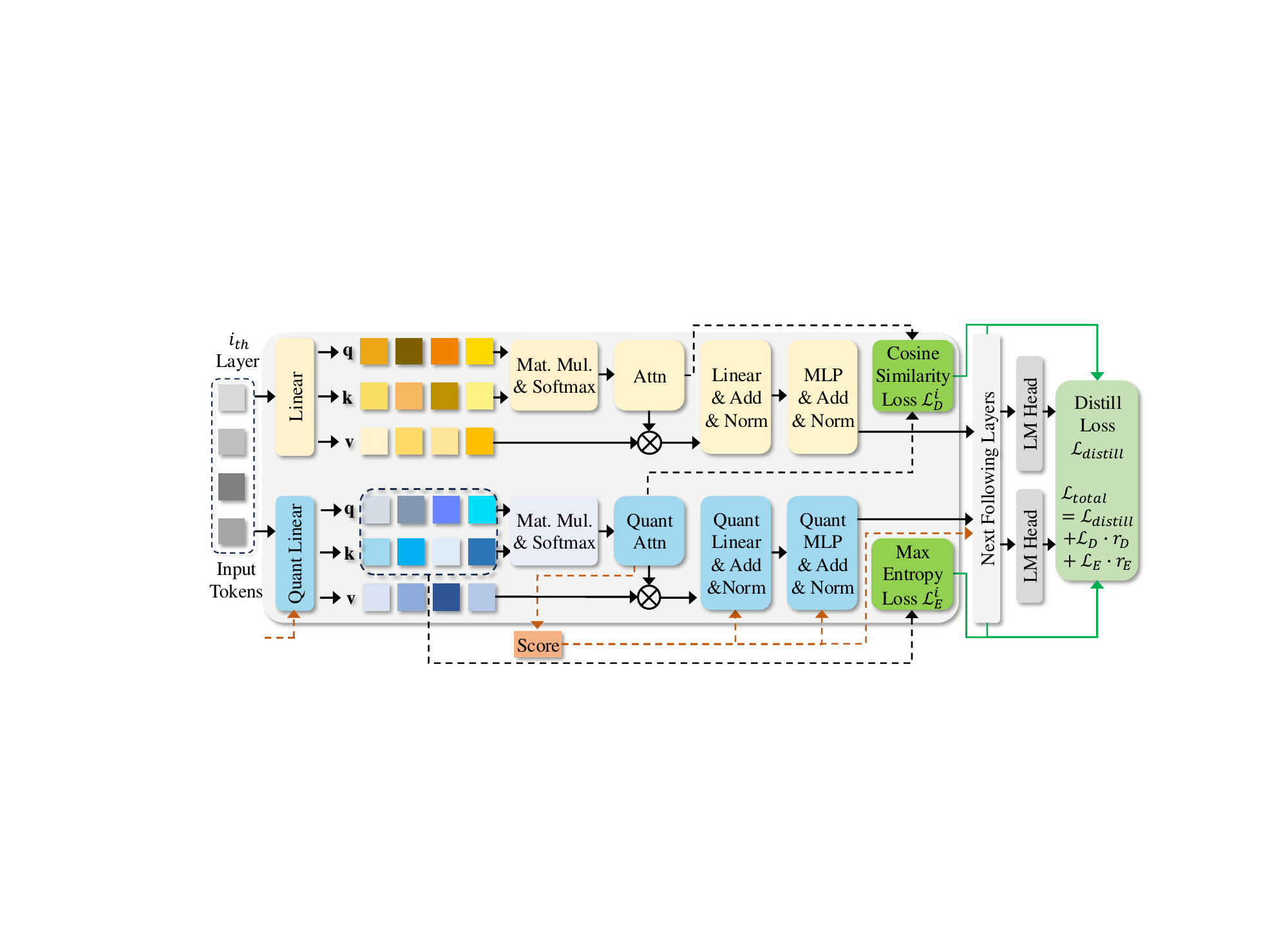}
  \caption{
  Adaptive distillation pipeline based on token importance score (colored in red), with maximum entropy loss and attention map cosine similarity loss (both colored in green).
  }
  \label{fig:training_pipeline}
\end{figure*}

\subsection{Entropy-Guided and Distribution-Aligned Optimization}

Based on the analysis in Section \ref{sec:motivations},  the performance loss is primarily attributed to the quantized attention module (especially the query and key) with deteriorated representation capability. 
To address this issue, we propose the entropy-guided and distribution-aligned optimization method, which statistically maximizes the entropy of representations and restores the capability of the quantized self-attention module.
According to the work~\cite{quant_entropy}, for Gaussian distributions, quantizers with maximum output entropy (MOE) and minimum average error (MAE) are approximately equivalent, up to a multiplicative constant.
In essence, minimizing the error caused by quantization is equivalent to maximizing the information entropy of quantized values.
As observed in Figure~\ref{fig:quant_query_key_distribution}, the distributions of the query $\textbf q$ and the key $\textbf k$ in the self-attention modules follow the Gaussian distribution as below,
\begin{equation} 
 \textbf{q} \sim \mathcal{N}(\mu_{\textbf{q}}, \sigma_{\textbf{q})}, 
\end{equation}
\begin{equation} 
    \textbf{k} \sim \mathcal{N}(\mu_{\textbf{k}}, \sigma_{\textbf{k}}).
\end{equation}
The entropy can be represented as follows, 
\begin{equation} 
    \mathcal{H}(\textbf{q}) = - \sum_{i} p(\textbf q_i) \log{p(\textbf q_i)} = \frac{1}{2} \log{2\pi e \sigma_{\textbf{q}}^{2}},
\end{equation}
\begin{equation} 
    \mathcal{H}(\textbf{k}) = - \sum_{i} p(\textbf k_i) \log{p(\textbf k_i)} = \frac{1}{2} \log{2\pi e \sigma_{\textbf{k}}^{2}}.
\end{equation}
To maximize the entropy $\mathcal{H}(\textbf{q}) \propto {\sigma_{\textbf{q}}^{2}}$ and $\mathcal{H}(\textbf{k}) \propto {\sigma_{\textbf{k}}^{2}}$  during the training process, we incorporate the entropy loss $ \mathcal{L}_{E}$ to optimize the total entropy of query and key for all layers and heads.
Specifically, we re-scale the entropy loss as follows:
\begin{equation} 
    \mathcal{L}_{E} = - \log\left(
     \sum_{l=1}^{L} \sum_{h=1}^{H} \log \left( 1+
    \sigma_{\textbf{q}}^{2} \sigma_{\textbf{k}}^{2}
    \right)
    \right),
\end{equation}
where $L$ and $H$ denote the number of layers and heads, respectively.
To prevent the occurrence of NaNs when scaling loss with log operation, we increment deviation product by 1.

Next, we focus on fixing the distribution pattern issue in the attention map.
As shown in Figure~\ref{fig:quant_quantized_fp16_attentionmap}, 
the column distribution pattern with the initial tokens from the FP16 counterpart disappears after quantization in the quantized attention map. 
To minimize the difference between the quantized attention map and the FP16 counterpart, 
a distribution loss  $\mathcal{L}_{D}$ is introduced based on the cosine similarity between the FP16 attention map ${attn}_f$ and quantized one ${attn}_q$ in each layer as follows:
\begin{equation} 
    \mathcal{L}_{D}  = \log \left( \sum_{l=1}^{L}  \sum_{h=1}^{H} \frac{{attn}_q \cdot {attn}_f}{ {\| {attn}_q \|}_2 \cdot {\| {attn}_f \|}_2 } \right).
\end{equation}
We re-scale the loss with the logarithmic operation to match the scale of the original loss.

\subsection{Token Adaptive Quantization}

Similar to token pruning~\cite{dong2023heatvit,wang2021spatten}, two features of the transformer structure: token-level redundancy and sequential computing, open up possibilities for token adaptive mixed-precision quantization.
Based on the analysis section, we assess the token importance with the averaged attentivity to the initial token through all transformer heads, denoted by the first column of the attention map (i.e., $attn[:,0]$).
Considering the trade-off between task performance and practical hardware efficiency, we assign 8 bits for more important tokens and 4 bits for less attentive ones.
Specifically, we adopt adaptive quantization for $\forall i \in [0,N-1]$ as follows,
\begin{equation} \label{eq:adaptive} 
    \beta( \textbf{x}_i \ | \ attn, \rho ) =
    \{
             \begin{array}{l}
             8, \ \ attn[i,0] \geq \text{TopK}(attn[:,0], \text{Int}(\rho * N)),  \\
             4, \ \ \text{others},
             \end{array} 
\end{equation}
where $\textbf{x}_i$ denotes $i^{th}$ token during training and generation processes, $\rho$ represents important token ratio, $N$ denotes number of tokens, function $\beta(\textbf{x}_i \ | \ attn, \rho)$ returns bit width for the $i^{th}$ token given attention map $attn$ and $\rho$, 
and $\text{TopK}(\cdot, k)$ denotes top-k function that returns $k_{th}$ largest element.

\begin{figure*}[t]
  \centering
  \includegraphics[width=1.0\linewidth]{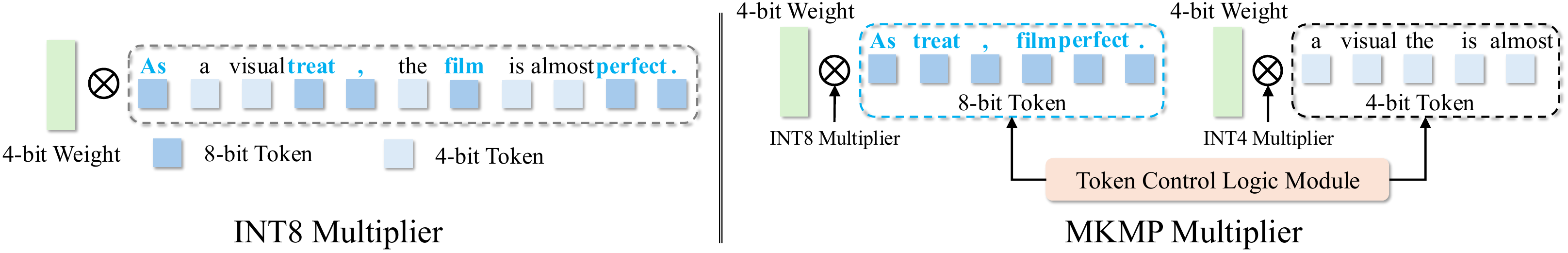}
  \caption{
  Comparison of INT8 multiplier and SIMD-based MKMP multiplier to support mixed-precision MAC for adaptive token quantization.
  }
  \label{fig:workflow_adaptive}
\end{figure*}


We design a Token Control Logic Module (TCLM) for adaptive quantization as shown in Figure~\ref{fig:workflow_adaptive}.
First, $\beta(\textbf{x}_i \ | \ attn[:,0], \rho)$ evaluates the importance of the $i^{th}$ input token according to the averaged attentivity.
When $\textbf{x}_i$ is informative, they are concatenated together for the following 8-bit layer-wise  integer quantization;
Otherwise, if $\textbf{x}_i$ is less informative, they are concatenated for the following 4-bit quantization.
After the layer-wise integer quantization,  our proposed MKMP multiplier is called to execute mixed integer MAC.
For the $\text{TopK}(\cdot, k)$ implementation, the fast top-k sorting operator, Heapsort, is leveraged, to support the $\rho$ important token selection.
Heapsort and Concatenation are existing operands with marginal
overhead.

\begin{figure}[t]
    \centering
    \includegraphics[width=\linewidth]{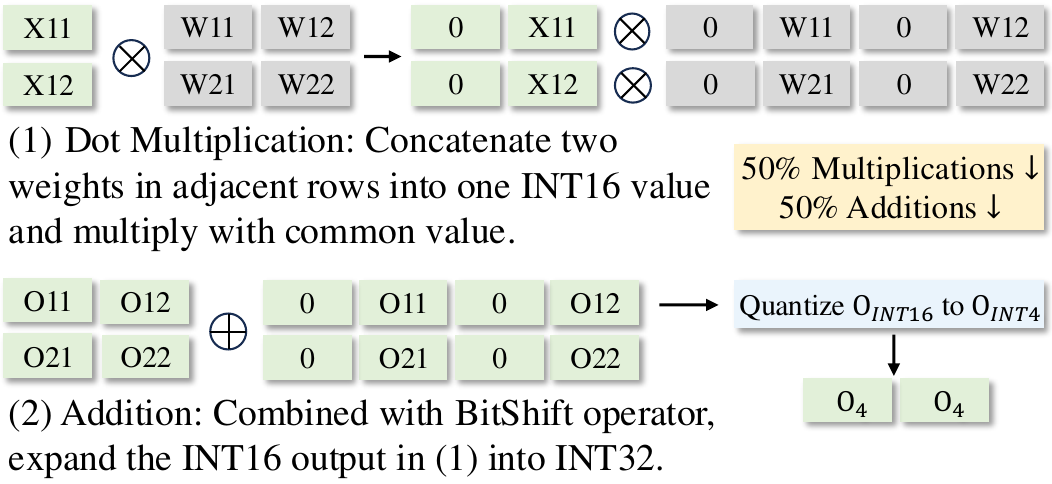}
\caption{SIMD-based INT4-concatenated multiplier design.}
\label{fig:multiplier_4bit}
\end{figure}

\subsection{Adaptive Training Pipeline}
We visualize our training pipeline in Figure~\ref{fig:training_pipeline}.
We use the FP16 model (colored in yellow) to distill the quantized model (colored in blue) during QAT. We apply soft distillation, which trains a student model to mimic a teacher model by minimizing the KL divergence between their softmax outputs~\cite{hinton2015distilling}.  
The distillation loss is defined as:
\begin{equation} 
\mathcal L_{distill} = (1- \gamma)\cdot\mathcal L_{CE}+ \gamma \tau ^2 \cdot \mathcal L_{KL},
\label{eq:soft_dis}
\end{equation}
where $\tau$ is the temperature for the distillation, and $\gamma$ is the coefficient balancing the KL divergence loss $\mathcal L_{KL}$ and the cross-entropy loss $\mathcal L_{CE}$.
In the quantization modules, the tokens are adaptively quantized with either 8 bits or 4 bits based on their scores (colored in red in Figure~\ref{fig:training_pipeline}) generated from the most recent attention map.

The entropy loss $\mathcal{L}_E$ and the distribution loss  $\mathcal{L}_{D}$ (both colored in green) are added to the total loss for optimization during training as follows,
\begin{equation} 
    \mathcal{L}_{total} = \mathcal{L}_{distill} + r_{E}  \cdot \mathcal{L}_{E}   +   r_{D}  \cdot 
 \mathcal{L}_{D},
\end{equation}
where the ratios $r_{E}$ and $r_{D}$ are used to scale the entropy and distribution losses, respectively.
In our experiments, we set $r_{E}=0.5$ and $r_{D}=1$
to facilitate better optimization.

\input{tables/blimp}

\subsection{Multi-Kernel Mixed-Precision Multiplier}

The standard SIMD-based INT8 multipliers do not support mixed-precision integer MAC operations and typically zero-extend sub-8-bit operands to byte boundaries as 8-bit operands.
To implement the proposed layer-wise token-adaptive quantization, we develop a SIMD-based MKMP multiplier to enable this mixed-precision quantization on devices as shown in Figure~\ref{fig:workflow_adaptive}.
After token adaptive quantization, we use existing INT8 multipliers for the 8-bit concatenated tokens and implement the INT4 multipliers.  
The INT4 multiplier is built on the existing INT8 multiplier, which concatenates weights from adjacent rows and multiplies them with a shared activation value in the SIMD kernel.

By concatenating weight matrices within GeMM, our approach significantly reduces the number of mathematical computation instructions required on processors compared to traditional byte-level quantized implementation kernels.
This reduction is proportional to the concatenation density for the same workload.
As shown in Figure~\ref{fig:multiplier_4bit}, two 4-bit operands are concatenated into a single 16-bit register unit.
This choice aligns with current practices in 8-bit quantized implementation kernels, where 8-bit data is often extended to 16 bits during product computations. This enables the use of efficient instructions like \textit{mla} in Arm ISAs, which utilize a 32-bit destination register (INT32 datatype) to perform multiplication and accumulation in a single instruction. The 16-bit intermediate registers facilitate concatenation while offering redundancy beyond the actual sub-byte values.
A low-bit priority strategy ensures that the bit width is utilized evenly, minimizing redundant zeros for subsequent computations.
The 16-bit-wide multiplication operation is then performed, with the results internally split to maintain mathematical accuracy for the subsequent addition steps. Theoretically, this design reduces the computational burden—both in terms of multiplications and additions—for 4-bit GeMM by half compared to the conventional approach, which expands 4-bit data to 8-bit for use in byte-level quantization kernels.

Notably, while this methodology originates from weight-matrix concatenation, it is equally applicable to activation-activation matrix multiplications in transformer models. Similar to weight matrices, one of the activation matrices can be concatenated while preserving the logic and characteristics of design. This versatility makes it a broadly adaptable low-bit acceleration strategy.
Using the SIMD-based memory mechanism, the INT4 multiplier employs bit-shift and row-by-row summation to add up intermediate values. INT4 multiplier can save 50\% hardware resources of INT8 multiplier.
By integrating quantization operator, we streamline entire MKMP multiplier within the GeMM kernel.
Due to LLMs' huge memory readout, we optimize and assign computing threads for different operations and overlap memory readout time from the compiler level.

%% file: tables/blimp.tex
\begin{table*}[t]
\centering
\caption{LLaMA-58M quantization results on the BLiMP dataset, including the BLiMP Supplement.}
\resizebox{1.03\linewidth}{!}{
\begin{tabular}{c|c|cccc|cccc|cccc}
\toprule
\# Bits        & FP16      & \multicolumn{4}{c|}{W8A8}                                                                                 & \multicolumn{4}{c|}{W4A8}                                                                                 & \multicolumn{4}{c}{W4A4}                                                                                 \\ \cmidrule{1-14}
Method      &  /         & \multicolumn{1}{c}{NIPQ} & \multicolumn{1}{c}{PACT} & \multicolumn{1}{c}{LLM-QAT} & \multicolumn{1}{c|}{Ours} & \multicolumn{1}{c}{NIPQ} & \multicolumn{1}{c}{PACT} & \multicolumn{1}{c}{LLM-QAT} & \multicolumn{1}{c|}{Ours} & \multicolumn{1}{c}{NIPQ} & \multicolumn{1}{c}{PACT} & \multicolumn{1}{c}{LLM-QAT} & \multicolumn{1}{c}{Ours} \\ \midrule
\multicolumn{14}{c}{BLiMP Main} \\ \midrule
AA       &  89.8   &    85.5                      &   86.4                     &        88.0                 &    {88.1}                      &        58.1                &   86.6                  &      87.1                   &       {87.6}                   &       66.2             &  {85.8}                  &      85.9                   &      85.7                    \\
AS   &    73.1    &        70.9                  &   70.7                    &         {72.4}                &    72.2                      &      55.5                    &    70.3                  &        72.4                 &     {72.3}                     &        54.4              & 69.6                  &        72.0                &   {71.3}                       \\
Bind.       &   72.7    &     71.1                     &71.0                      &        71.9                 &     {72.3}                     &         61.7                 &    70.6                   &        71.8                 &     {72.2}                     &     51.5                   &  68.2                 &       71.5                  &   {72.4}                       \\
C/R    &  67.5      &   65.5                       &  64.6                      &       66.6                  &     {66.7}                     &      54.7                    &    64.0                    &         65.8                &                 {66.7}         &     53.6                    &  63.6                     &           65.4              &                {66.3}          \\
D-NA    &   90.8     &    86.9                      &     86.3               &           89.0              &     {89.2}                     &        54.2                  &    86.6                     &          90.1              &                {89.1}          &      53.4                  &  84.8                    &      87.1                   &               {87.5}           \\
Ell.       &  73.3    &    60.4                      &  59.7              &       68.4                  &     {69.4}                     &           29.9               &   59.7                 &           67.2              &    {69.8}                      &           33.8             &  56.8                   &       63.2                  &    {65.1}                      \\
F-G    &    71.8   &  70.2                        &  69.0              &           71.8              &      {72.1}                    &         66.7                 &  69.3                  &        71.7                 &    {72.0}                      &       61.1               &  66.8                  &         70.2                &   {70.4}                       \\
IF    &  93.1     &                 94.6         &  94.8                    &        {95.1}                 &                 95.0         &     45.8                     &   {95.2}                   &          93.3               &                94.9          &    52.2                     &  93.7                    &      94.1                   &               {94.9}           \\
IE   &   51.2    &                48.2          &  49.2               &    51.3                    &                      {51.7}    &         43.6                 &   50.0                   &      51.9                   &                   {52.1}       &        48.5                &   43.3                 &         48.2                &                 {51.3}         \\
NPI-L    &    56.5     &                50.0          &   52.1               &               57.9          &                   {58.3}       &         26.8                 &   52.2                   &           57.3              &               {57.7}           &     36.6                   &  48.2                   &          {45.9}               &              44.5            \\
Quan.      &   73.3      &                  73.7        &  75.8               &            {81.0}             &                     79.0     &        57.2                  &   78.2                 &           {79.4}              &                    79.3      &      42.7                 &  78.0                     &          78.2              &                 {80.0}         \\
S-VA    &  75.4     &                  68.4        &   67.8             &          73.1               &                    {73.2}      &           46.3               &   67.7                   &           73.0              &                   {74.0}       &       48.6                &  64.5                    &         68.0                &                   {70.3}       \\ \midrule
Avg.   &   74.0    &                 70.5        &   70.6             &          73.8               &                    \textbf{73.9}      &           50.0               &   70.9                   &           73.4              &                   \textbf{74.0}       &       50.2                &  68.6                   &         71.0                &                   \textbf{71.8}       \\ \midrule
\multicolumn{14}{c}{BLiMP Supplement} \\ \midrule
Hyper.     &   49.3     &                 48.0         &   49.0             &            {49.6}             &                   48.9       &       49.5                   &  48.7                   &        48.7                 &                 {49.6}         &     {50.9}                    &  50.3                   &     49.3                    &                   50.5       \\
QAC-E     &   51.6      &             48.4             &   {51.5}                 &         49.1                &                50.1          &       35.9                   &  50.0                      &      49.8                   &           {50.1}               &     37.5                    &  48.4                     &           49.3              &            {50.1}              \\
QAC-t    &   41.8    &              40.6            &  40.0               &       {41.6}                  &                 41.3         &       34.5                   &   40.6                   &          40.6               &              {41.3}           &   33.9                      &  39.3                     &        40.6                 &             {41.9}             \\
S-AI    &    88.5    &               {89.1}           &  87.9                   &            88.6             &                88.5          &       67.8                   &  {89.8}                  &       89.1                  &              89.2          &    54.6                     &  87.3                      &        87.3                 &             {89.0}             \\
TT     &   66.1   &                     58.2     &   57.1             &          {62.0}               &                      61.5    &        43.2                  &  57.5              &     60.3                    &                   {61.8}       &      51.4                  &   55.7                 &       59.2                  &                  {60.1}        \\ \midrule
All Avg.   &    69.7    &                     66.5     &   66.6             &          69.2               &      \textbf{69.3}                    &        48.9                  &  66.9              &     68.8                    &     \textbf{69.4}                    &      48.9                  &   64.9                 &       66.9                  &   \textbf{67.8}                       \\ \bottomrule
\end{tabular}
}
\label{tab:blimp_main_results_table}
\end{table*}


%% file: sections/5_results.tex
\section{Experimental Setup}

\subsection{Quantization Setup}
For the verification and deployment of our proposed methods,
we experiment with lightweight LLMs, 
including LLaMA-58M~\cite{touvron2023llama, timiryasov2023babyllama} and GPT2-97M~\cite{gpt2}.
We adopt the pretrain datasets from the work~\cite{babyllama_challenge} and then perform regex-based cleaning on them.
The cleaned datasets are tokenized using BytePair Encoding (BPE) with a vocabulary size of 16000.
The models are then evaluated on BLiMP~\cite{blimp} for the zero-shot test,
and (Super) GLUE~\cite{wang2019superglue} for the fine-tuning test.
In the absence of prior coarse-grained QAT studies for LLMs, we compare with well-known static quantization methods as baselines, including NIPQ~\cite{park2022nipq}, PACT~\cite{choi2018pact}, and LLM-QAT~\cite{liu2023llmqat}.
The same fine-tuning recipe with distillation based on the FP16 pretrained model is adopted for all experiments.

\subsection{Hardware Deployment}

We use the OnePlus 11 smartphone, powered by the Snapdragon 8 Gen 2, as our mobile platform, utilizing all available cores for multi-threaded computation. Similarly, on the Raspberry Pi 5 with its BCM2712 quad-core Arm Cortex A76 processor, we deploy our quantized model and distribute the computations across all four cores. Latency is reported based on 1000 iterations for each test.

\section{Experimental Results}

\subsection{Zero-Shot Evaluation}

We first verify the effectiveness of our proposed QAT framework on the BLiMP~\cite{blimp} dataset with zero-shot (i.e., no fine-tuning) evaluations, and the results are shown in Table~\ref{tab:blimp_main_results_table}.
We compare our method with the other three QAT works, including NIPQ, PACT, and LLM-QAT, under different bit-width settings including W8A8 (meaning 8-bit weight and 8-bit activation quantization), W4A8, and W4A4.
As observed, our approach achieves better performance than all other three works in terms of the average accuracy of all subdatasets on the BLiMP dataset.
Our method performs the best  on most of the subdatasets across three bit-width configurations.
Especially for the W4A8 setting, which is the most  practical in wide applications, our method achieves an average accuracy of 69.4\%, which is close to that of the FP16 model (only 0.3\% drop) and even surpasses the W8A8 setting (69.3\%).
For the W4A4 setting, our method maintains an average accuracy of 67.8\%, showcasing a clear advantage over other methods.
Only our method can achieve a competitive average accuracy close to that of the FP16 model, while the baselines usually suffer from substantial accuracy drops. 
NIPQ fails to restore the accuracy when the model weights are quantized to 4 bits.
For PACT, it is sensitive to the bit width of the activations, as evidenced by the poor results under the W4A4 setting.
The LLM-QAT method consistently produces models with an lower average accuracy than our method.

\subsection{Generalization Verification}
Additionally, we deliver the evaluation results of the GPT2-97M model with the W4A4 setting to verify the generalization of our method in Table~\ref{tab:gpt2_results_table}.
We conduct the experiments on the BLiMP main dataset.
Our method can achieve the highest average accuracy with the best performance on most of the subdatasets, demonstrating our clear advantages over QAT baselines.
Among the baselines struggling to restore the  accuracy, the NIPQ and PACT perform much worse with large margins. 
Thus, the clear advantages, achieved by our method compared to other QAT methods, validates the generalization of our proposed Squant method for the small language models.

\section{Fine-Tuning Evaluation}
To further demonstrate the effectiveness of the proposed Squant framework, we finetune the quantized models from different QAT frameworks on the (Super) GLUE dataset and show the evaluation results in Table~\ref{tab:superglue_results_table}.
To make a fair comparison, we use the same finetuning recipe for all methods.
As observed, the proposed Squat method can restore the performance on all subdatasets and demonstrate a clear advantage in average accuracy compared to all the other three methods.
In detail, other methods struggle to optimize the quantized model.
The NIPQ can only restore the model performance on the WSC subdataset, and fail to the average accuracy.
The PACT and LLM-QAT methods yield poor results on some subdataset.
For instance, PACT exhibits bad results on the RTE and MultiRC subdatasets, while LLM-QAT experiences significant performance losses on the WSC subdataset.
Therefore, the effectiveness of our proposed Squant framework on the down-streaming tasks is verified by the clear accuracy advantages.


\input{tables/gpt2-blimp}

\input{tables/llama-58m-superglue}

\section{Hardware Efficiency}

Our MKMP multiplier is compatible with mainstream processors on edge platforms, such as mobile phones and Raspberry Pi IoT processors, which typically face challenges when processing low-bit data due to their SIMD instructions supporting only 8-bit or larger data granularity.
We deliver the latency profiling results with model size and accuracy in Table~\ref{tab:hardware_results}, and we can draw the following conclusions: 8-bit quantization provides more than 1.4$\times$ acceleration on smartphones and over 1.6$\times$ acceleration on Raspberry Pi. 
As the high-end CPUs on smartphones can afford more robust floating-point processing capabilities, the acceleration attained through quantization on smartphones is not as significant as the improvements observed on the Raspberry Pi 5.
Meanwhile, for the W4A4 configuration, we achieve more than 2.2$\times$ acceleration on smartphones and 2.3$\times$ acceleration on Raspberry Pi, separately.
Overall, the GPT2-97M model achieves greater acceleration in our framework compared to the LLaMA-58M model.
This is largely due to its higher parameter amount, which enables more efficiency improvement through memory access reduction on edge devices.
Additionally, the 4-bit compression and concatenation technique amplifies this advantage, delivering a 2.26$\times$ acceleration compared to the 1.43$\times$ speedup achieved with 8-bit quantization on smartphones for GPT2-97M model.

Also, the introduction of mixed precision is essential as it bridges the gap between the latency of 4-bit and 8-bit configurations.
While the theoretical, computational workload is halved, some overhead is introduced due to internal shifts of concatenated weights and the recovery of stored results in INT8 format. However, using W4A4 precision can lead to a noticeable performance drop in LLM tasks. To address this, we employ our MKMP multiplier for the mixed W4A4 and W4A8 configurations.
This approach not only achieves further acceleration compared to 8-bit quantization but also maintains the model performance as shown in Figure~\ref{fig:mix_uniform_quant_ablatioin}. 
For Raspberry Pi, the additional acceleration achieved through the mixed configuration becomes over 40\%. 

\input{tables/hardware}

\section{Ablation Study}

\subsection{Loss Ablation}
As shown in Figure~\ref{fig:ablation_loss}, we adopt ablation study for proposed entropy loss $\mathcal{L}_{E}$ and distribution loss $\mathcal{L}_{D}$.
The results in blue denote the LLaMA-58M and the results in red denote GPT2-97M.
The results are obtained with the W4A4 configuration.
We can identify that, compared to entropy loss, distribution loss more effectively improves the model performance.
Besides, we make the observation that combining the two losses generates better results than using either single one.
Both loss types are verified to be effective when used for both LLaMA-58M and GPT2-97M, which validates the generalization of the proposed loss optimization method.

\subsection{Mixed Strategy}
Ablation for quantization with mixed or uniform strategy is included in Figure~\ref{fig:mix_uniform_quant_ablatioin}.
The results in blue denote the quantization with mixed strategy while the results in grey denote the quantization with uniform strategy.
Results are evaluated with LLaMA-58M model on BLiMP Main dataset using a Raspberry Pi 5.
Results show that mixed strategy yields superior quantization performance (higher accuracy and lower latency in ms/Token) compared to uniform strategy.
The inference acceleration using a mixed strategy verifies superior performance compared to uniform quantization at any bit level (5, 6, 7, or 8 bits).
Specifically, for 6-bit activation quantization strategy, the mixed-precision strategy with half 4-bit and half 8-bit quantization shows impressive better accuracy than the uniform strategy.



%% file: tables/gpt2-blimp.tex
\begin{table}[t]
\centering
\caption{GPT2-97M with W4A4 on BLiMP Main dataset.}
\resizebox{1.0\linewidth}{!}{
\begin{tabular}{c|c|cccc}
\toprule
Method          & FP16 & NIPQ & PACT & LLM-QAT & Ours \\
\midrule
AA  & 87.0 & 38.1 & 69.8 & 84.3 & {84.5} \\
AS  & 71.3 & 57.4 & 63.7 & 70.5 & {71.7} \\
Bind.         & 70.2 & 49.8 & 64.4 & 69.7 & {69.8} \\
C/R & 66.1 & 54.2 & 62.6 & 65.1 & {65.3} \\
D-NA  & 87.4 & 51.4 & 72.3 & {86.9} & 86.0 \\
Ell.        & 62.1 & 39.6 & 39.2 & 59.8 & {59.9} \\
F-G      & 70.7 & 43.3 & 63.2 & 70.5 & {70.4} \\
IF & 94.1 & 52.3 & 90.0 & 94.3 & {95.4} \\
IE  & 47.2 & {59.7} & 44.9 & 46.5 & 46.8 \\
NPI-L  & 48.5 & {71.3} & 44.4 & 47.5 & 44.8 \\
Quan.     & 68.0 & 27.5 & 46.7 & 69.5 & {69.4} \\
S-VA. & 66.2 & 48.1 & 55.5 & 65.1 & {66.0} \\
\midrule
Avg.         & 69.9 & 49.4 & 59.7 & 69.1 & \textbf{69.2} \\
\bottomrule
\end{tabular}
}
\label{tab:gpt2_results_table}
\end{table}

%% file: tables/llama-58m-superglue.tex
\begin{table}[]
\centering
\caption{
LLaMA-58M with W4A4 on (Super)GLUE.
}
\resizebox{1.0\linewidth}{!}{
\begin{tabular}{c|c|cccc}
\toprule
Method  & FP16 & NIPQ & PACT & LLM-QAT & Ours \\
\midrule
CoLA    &   69.5   &   33.3   &   {69.3}   &  68.5   &   68.4   \\
SST-2   &   87.2   &   49.4   &   {85.4}   &  85.0   &   84.1   \\
MRPC    &   63.2   &   32.2   &   69.4   &  69.3   &   {69.5}   \\
QQP     &   84.3   &   42.4   &   82.5   &  83.7   &   {84.1}   \\
MNLI    &   72.9   &   35.4   &   67.5   &  70.8   &   {70.8}   \\
MNLIm &   73.7   &   35.8   &   69.1   &  {71.5}   &   71.1   \\
QNLI    &   81.1   &   47.2   &   74.4   &  78.2   &   {79.4}   \\
RTE     &   61.6   &   50.5   &   48.5   &  {54.6}   &   53.5   \\
BoolQ   &   67.2   &   58.4   &   60.3   &  {62.4}   &   62.9   \\
MulRC &   58.9   &   53.2   &   46.1   &  53.7   &   {54.1}   \\
WSC     &   61.4   &   {61.4}   &   53.0   &  52.9   &   56.6   \\
\midrule
Avg.    &   71.0   &   45.4   &   65.9   &  68.2   &   \textbf{68.6}   \\
\bottomrule
\end{tabular}
}
\label{tab:superglue_results_table}
\end{table}

%% file: tables/hardware.tex
\begin{table*}[t]
\centering
\caption{
Latency results (ms/Token) of LLaMA-58M and GPT2-97M with 128 input sequence length on mobile (Onepluss 11) and edge (Raspberry Pi 5) devices.
}
\resizebox{1.0\linewidth}{!}{
\begin{tabular}{c|c|cc|ccc|c}
\toprule
  W      & FP16 & INT8 & INT4 & INT4          & INT4          & INT4         & INT4 \\
  A      & FP16 & INT8 & INT8 & 4:8 (1:3) & 4:8 (1:1) & 4:8 (3:1) & INT4 \\
\toprule
 \multicolumn{8}{c}{LLaMA-58M  (ms/Token)}   \\
\bottomrule
 MB        &   110.6   & 55.3   & 27.7  &   27.7         &      27.7      &   27.7           & 27.7  \\
\midrule
 Mobile  &   4.54   & 3.22 (1.41$\times$)  & 2.56 (1.77$\times$)     &  2.39 (1.90$\times$)  & 2.23 (2.04$\times$) & 2.10 (2.16$\times$)  &  2.02 (2.24$\times$) \\
Raspberry Pi &  15.63     & 9.40 (1.66$\times$)  &7.50 (2.08$\times$)  & 7.30 (2.14$\times$)  & 7.08 (2.21$\times$) & 6.89 (2.27$\times$)   & 6.78 (2.31$\times$) \\
\toprule
\multicolumn{8}{c}{GPT2-97M (ms/Token)}   \\
\bottomrule
MB       &   185.5   &  92.7 &  46.3 &      46.3      &      46.3      &   46.3           &  46.3 \\
\midrule
Mobile     &   6.22    &  4.35 (1.43$\times$) & 3.42 (1.82$\times$)   & 3.06 (2.06$\times$)    & 3.02 (2.03$\times$)   &  2.86 (2.17$\times$)    &  2.75 (2.26$\times$) \\
Raspberry Pi &   23.04    & 13.75 (1.68$\times$)  & 12.45  (1.85$\times$)   &  11.24  (2.05$\times$)  & 10.98  (2.10$\times$)     &  10.01  (2.30$\times$)  & 9.74 (2.37$\times$) \\
\bottomrule
\end{tabular}
}
\label{tab:hardware_results}
\end{table*}

%% file: sections/6_conclusion.tex
\section{Conclusion and Limitation}

In this paper, we introduce the Squant method, an entropy-guided and distribution-aligned token adaptive mixed-precision QAT framework, designed to accelerate small language models on mobile devices.
Besides, we adaptively quantize tokens with different bit widths based on their importance, which further accelerates the inference and maintains performance.
Meanwhile, we implement the corresponding multiplier, which helps the mobile devices benefit from our proposed 4-bit quantization optimization algorithm.
We effectively restore the model performance to that of FP16 counterparts and achieve up to 2.37$\times$ speedup on mobile devices.
We will verify our method on larger models with hundreds of millions of parameters  in our further work. 

\begin{figure}[]
    \centering
    \includegraphics[width=\linewidth]{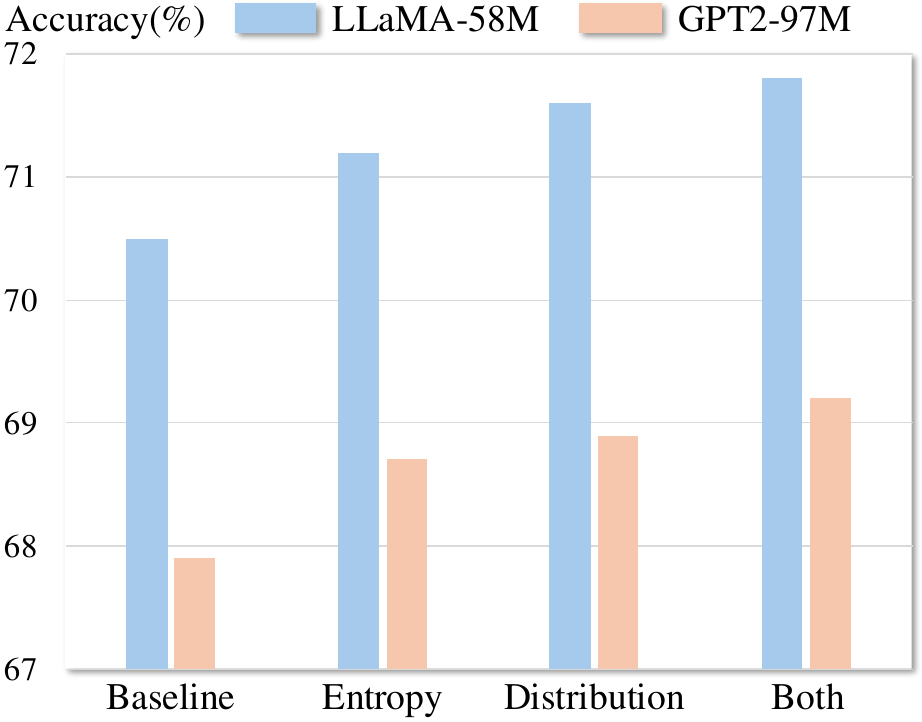}
    \caption{
    Loss ablation with average accuracy of LLaMA-58M and GPT2-97M on BLiMP Main dataset in W4A4.
    }\label{fig:ablation_loss}
\end{figure}

\begin{figure}[]
    \centering
    \includegraphics[width=\linewidth]{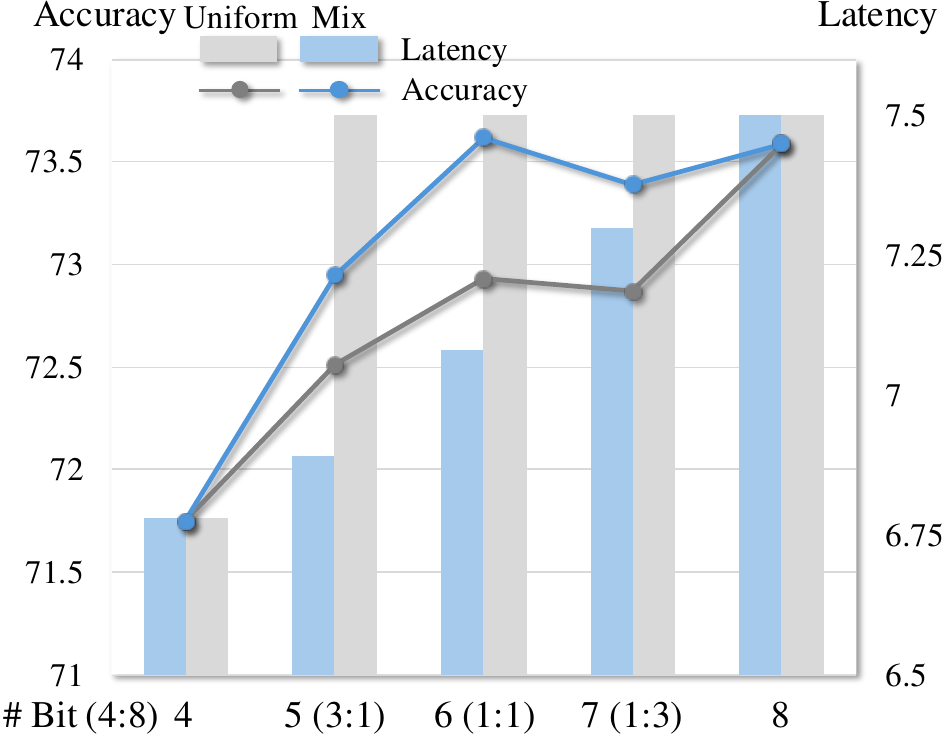}
    \caption{
    Mixed and uniform quantization results of LLaMA-58M on the BLiMP Main dataset with Rapberry Pi 5.
    }\label{fig:mix_uniform_quant_ablatioin}
\end{figure}